\pdfoutput=1
\documentclass[11pt]{article}

\usepackage[final]{acl}
\usepackage{times}
\usepackage{latexsym}
\usepackage[T1]{fontenc}
\usepackage[utf8]{inputenc}
\usepackage{microtype}
\usepackage{inconsolata}
\usepackage{booktabs}
\usepackage{graphicx}
\usepackage{amsmath}

\usepackage{amssymb}
\usepackage{tabularx}
\usepackage{standalone}  
\usepackage{xcolor}

\usepackage{float}
\definecolor{cloudburstColor}{RGB}{0,114,178}    
\definecolor{composerColor}{RGB}{213,94,0}         
\definecolor{citizenshipColor}{RGB}{34,139,34}      

\usepackage{tikz}
\usetikzlibrary{calc} 
\usetikzlibrary{arrows.meta, positioning}
\tikzset{
  ent/.style={rectangle, draw, align=center, minimum width=2.5cm, minimum height=1cm, font=\Large}
}
\title{Multi-Hop Reasoning for Question Answering with Hyperbolic Representations}

\author{\textbf{Simon Welz}$^{1}$ \quad \textbf{Lucie Flek}$^{1,2}$ \quad \textbf{Akbar Karimi}$^{1,2}$ \\ 
$^1$Bonn-Aachen International Center for Information Technology, University of Bonn, Germany \\
$^2$Lamarr Institute for ML and AI, Germany \\
\texttt{ak@bit.uni-bonn.de}
}

\begin{document}

\maketitle

\begin{abstract}
Hyperbolic representations are effective in modeling knowledge graph data which is prevalently used to facilitate multi-hop reasoning. However, a rigorous and detailed comparison of the two spaces for this task is lacking. In this paper, through a simple integration of hyperbolic representations with an encoder-decoder model, we perform a controlled and comprehensive set of experiments to compare the capacity of hyperbolic space versus Euclidean space in multi-hop reasoning. Our results show that the former consistently outperforms the latter across a diverse set of datasets. In addition, through an ablation study, we show that a learnable curvature initialized with the delta hyperbolicity of the utilized data yields superior results to random initializations. Furthermore, our findings suggest that hyperbolic representations can be significantly more advantageous when the datasets exhibit a more hierarchical structure.
\end{abstract}
\section{Introduction}

Multi-hop reasoning is a complex task that requires models to integrate information across multiple pieces of evidence to arrive at accurate conclusions. For instance, to answer \textit{Which country is the composer of the song Cloudburst from?} using a simple knowledge graph like Figure \ref{fig:sample_knowledge_graph}, the model has to first find the answer to the first relation (composer of Cloudburst), and then look for the answer to the second relation (Whitacare's country of citizenship). This inherently involves traversing hierarchical relationships, making it particularly challenging for traditional language models that rely on Euclidean representations. While Euclidean representations are commonly used and can capture hierarchical structures to some extent \cite{nguyen2023cyle, misra2023triggeringmultihopreasoningquestion, zhang2024end}, recent research has shown that hyperbolic representations are more effective in handling such data due to their superior ability to model hierarchical and relational information \cite{nickel2017poincare, dhingra2018embedding, balažević2019multirelationalpoincaregraphembeddings, chami2020lowdimensionalhyperbolicknowledgegraph, xu2022hyperminer}.
\begin{figure}[t]
\begin{center}
    \scalebox{0.55}{
      \begin{tikzpicture}[node distance=2cm]
          \node[ent, fill=cloudburstColor!20, text=cloudburstColor] (cloudburst) {Cloudburst};
          \node[ent] (whitacre) [below left=of cloudburst] {Eric Whitacre};
          \node[ent] (searle) [below right=of cloudburst] {Francis Searle};
          \node[ent] (american) [below=of whitacre] {American};
          \node[ent] (birth) [below=of searle, xshift=-2cm] {14 March 1909};
          \node[ent] (death) [below=of searle, xshift=2cm] {31 July 2002};

          \draw[->] (cloudburst) -- node[above, sloped, font=\Large, color=composerColor] {composer} (whitacre);
          \draw[->] (cloudburst) -- node[above, sloped, font=\Large] {director} (searle);
          \draw[->] (whitacre) -- node[right, font=\Large, color=citizenshipColor, yshift=7] {country of citizenship} (american);
          \draw[->] (searle) -- node[left, font=\Large] {date of birth} (birth);
          \draw[->] (searle) -- node[right, font=\Large] {date of death} (death);
      \end{tikzpicture}
    }
\end{center}

\small
\noindent \textbf{Question:} Which \textcolor{citizenshipColor}{country} is the \textcolor{composerColor}{composer} of the song \textcolor{cloudburstColor}{Cloudburst} from?\\[1em]
\noindent \textbf{Path:} \textcolor{cloudburstColor}{Cloudburst} $\rightarrow$ \textcolor{composerColor}{composer} $\rightarrow$ Eric Whitacre $\rightarrow$ \textcolor{citizenshipColor}{country of citizenship} $\rightarrow$ \textit{American}
\caption{A knowledge graph with entities as the nodes and the relations as the edges, illustrating a 2-hop question answering process.}
\label{fig:sample_knowledge_graph}
\end{figure}

In multi-hop reasoning tasks, hierarchical reasoning often manifests in navigating knowledge graphs or layered question answering frameworks, where the complexity increases with each additional hop. Given the hierarchical nature of multi-hop reasoning tasks, hyperbolic geometry presents a compelling alternative to Euclidean space for embedding representations. Specifically, hyperbolic space provides a larger capacity and more expressive way of encoding tree-like and graph-like structures \cite{chami2019hyperbolicgcnn}, which are prevalent in knowledge graphs used for reasoning.

Although many studies successfully implement hyperbolic architectures and report performance gains \cite{zhou2021multi, xiao2022hyperbolic, wang2024hypermr}, they often fail to disentangle the effects of geometric properties from the introduction of additional trainable parameters and the resulting model architecture. 
 
A controlled comparison incorporating equivalent Euclidean architectures with comparable parametric complexity would be necessary to isolate the geometric contribution to model performance. In addition, models in the literature often require significant architectural changes, which can increase model complexity and computational costs. In contrast, our approach focuses on incorporating hyperbolic geometry into existing language model architectures with minimal changes.

In this paper, \textbf{we address the lack of a carefully controlled comparative study for evaluating the differences between hyperbolic and Euclidean spaces in multi-hop reasoning}. As such, we incorporate hyperbolic representations into an encoder-decoder via the addition of a single layer and exponential mapping operation in the Poincaré ball model of hyperbolic space. We conduct a comprehensive set of experiments on multiple datasets, demonstrating that \textbf{adding a hyperbolic layer to increase the learning capability of a language model consistently outperforms its Euclidean counterpart}. In addition, we perform an ablation study to evaluate the impact of initialization of the curvature. Our results indicate that initializing the curvature using the $\delta$-hyperbolicity of the dataset leads to superior performance compared to random initialization. Furthermore, we show that the \textbf{performance gain from hyperbolic representations is more pronounced for datasets with more hierarchical structures} (defined based on the number of out-going relations for the nodes). Our comprehensive ablation studies deepen our understanding of geometric learning advantages in the context of language models and underscore the importance of aligning the geometric properties of the model with the inherent structure of the data.

\section{Related Work}

\textbf{Knowledge-based multi-hop reasoning.} Multi-hop reasoning requires traversing relational paths to synthesize new knowledge, making it essential for question answering. In the early approaches, path-based methods were utilized, in which reasoning was conducted using predefined rules or relational paths within the KB \cite{lao-etal-2011-random}.
Although these approaches were interpretable, they were frequently constrained by the availability and completeness of the KB. Neural-based reasoning models, including embedding-based methods \cite{bordes2013transe, wang2014knowledge, yang2015embedding, sunrotate}, introduced vectorized representations of entities and relations, enabling reasoning through learned relational patterns. More recent work has integrated graph-based neural architectures, such as Graph Convolutional Networks \cite{schlichtkrull2018modeling} and Graph Attention Networks \cite{velivckovic2018graph}, to propagate information across multi-hop relational structures in KBs. Reinforcement learning has been effectively applied to multi-hop reasoning over knowledge bases, enabling models to navigate complex relational paths and infer missing information. In this context, an RL agent is trained to traverse a KB by selecting a sequence of relations and entities, forming a reasoning path that leads to the desired answer \cite{ma2024pmhr, wan2021reasoning, lin-etal-2018-multi, zhu2022step}.

\noindent
\textbf{Hyperbolic multi-hop reasoning.} Recent studies have proposed frameworks that leverage hyperbolic geometry to enhance multi-hop reasoning capabilities \cite{zhou2021multi, xiao2022hyperbolic, wang2024hypermr}. Hyperbolic knowledge graph embeddings have demonstrated significant potential for multi-hop reasoning to model the hierarchical relationships in knowledge graphs \cite{chami2020lowdimensionalhyperbolicknowledgegraph, balažević2019multirelationalpoincaregraphembeddings, montella2021hyperbolictemporalknowledgegraph, kolyvakis2019hyperkg}. More recently, hyperbolic graph neural networks (HGNNs) have emerged as a promising direction for improving multi-hop reasoning \cite{liu2023hyperbolicgraphreasoning, liu2019hyperbolicgnn, chami2019hyperbolicgcnn}. These models extend traditional GNNs by incorporating hyperbolic message passing, allowing for better hierarchical aggregation of multi-hop dependencies. 

While these works highlight the potential of fully hyperbolic architectures, they often make significant architectural changes and fail to disentangle the advantages of hyperbolic geometry from those modifications in the model. By simply adding a hyperbolic layer to an existing language model, we efficiently integrate it without sacrificing the scalability or flexibility of these models. Furthermore, we utilize an identical model with the same number of trainable parameters to compare the two geometries, resulting in a deeper understanding of the characteristics of the two spaces.

\section{Background}

\subsection{Multi-Hop Reasoning}
Multi-hop reasoning can be defined as a process by which conclusions or answers are derived by sequentially combining information from multiple pieces of evidence. In contrast to single-hop reasoning, which relies on direct connections between a query and its answer, multi-hop reasoning involves traversing intermediate steps or relationships to reach the outcome. This capability is essential for tasks that require analyzing interconnected data or reasoning through layered information.
The process of multi-hop reasoning necessitates the retrieval of relevant pieces of information and their coherent integration, often involving the navigation of hierarchical relationships, temporal sequences, or contextual dependencies within data. A common approach to facilitate multi-hop reasoning is through the utilization of knowledge graphs, which represent entities and their relationships as a network. In a knowledge graph, reasoning involves following edges between entities to combine information across multiple nodes, thereby enabling complex inference over interconnected facts.

\subsection{Poincaré Ball Model}
As previous work has demonstrated the effectiveness of the Poincaré ball model \cite{nickel2017poincare, ganea2018hyperbolicneuralnetworks, chami2020lowdimensionalhyperbolicknowledgegraph, khrulkov2020hyperbolicimageembeddings, chen-etal-2024-hyperbolic} in capturing hierarchical relationships, we adopt it to enhance our ability to model such structures efficiently. The Poincaré ball provides a hyperbolic space where points are confined within the unit ball, enabling the representation of complex hierarchal structures with increasing precision near the boundary.
Similarly to the approach in \citet{nickel2017poincare}, we define the Poincaré ball model as $\mathbb{B}_c^n = \{x \in \mathbb{R}^n : c\| x \|^2 < 1, c\geq 0\}$. This space is equipped with a conformal factor given by: $\lambda_x^c = \frac{2}{1 - c\| x \|^2}$ where the hyperparameter 
$c$ determines the curvature of the space, with larger values of 
$c$ corresponding to spaces of higher negative curvature.

\noindent
\textbf{Möbius addition.} For a pair $x,y \in \mathbb{B}_c^n$, the Möbius addition is defined as follows:
\[
x \oplus_c y := \frac{(1+2c\langle x, y \rangle + c\| y \|^2)x + (1-c\| x \|^2)y}{1 + 2c\langle x, y \rangle + c^2\| x \|^2\| y \|^2}
\]
where $\langle x, y \rangle$ denotes the Euclidean inner product, and $\| \cdot \|$ represents the Euclidean norm.

\noindent
\textbf{Distance.} The induced distance function in this model is expressed as:
\[
d_c(x,y) := \frac{2}{\sqrt{c}} \arctan(\sqrt{c}\|-x \oplus_c y\|)
\]
which captures the geodesic distance between points $x$ and $y$ within the Poincaré ball.

To transition between Euclidean and hyperbolic spaces, we utilize the exponential and logarithmic mappings:

\noindent
\textbf{Exponential mapping} maps a Euclidean vector \( v \in T_o\mathbb{D} \) (the tangent space at the origin) to a point \( y \in \mathbb{D}_c^n \) on the Poincaré ball:  
    \[
    \exp_0^c(v) = \frac{\tanh\left(\|v\| \cdot \sqrt{c}\right)}{\|v\| \cdot \sqrt{c}} \cdot v
    \]

\noindent
\textbf{Logarithmic mapping} maps a point \( y \in \mathbb{B}_c^n \) back to the tangent space at the origin \( T_o\mathbb{B}_c^n \):  
    \[
    \log_0^c(y) = \frac{\tanh^{-1}(\|y\| \cdot \sqrt{c})}{\|y\| \cdot \sqrt{c}} \cdot y
    \]

\noindent
\textbf{Poincaré linear layer.}
The Poincaré linear layer, adapted from \citet{shimizu2021hyperbolicneuralnetworks, vanspengler2023poincareresnet}, extends the concept of a Euclidean linear layer into hyperbolic space, enabling models to effectively process hierarchical data. For an input $x \in \mathbb{D}_c^n$ the layer computes a hyperbolic transformation parameterized by weights $Z = \{z_k \in \mathbb{R}^n\}$ and biases $r = \{r_k \in \mathbb{R}\}_{k=1}^m$. The transformed output for each class $k$ is obtained through the following formulation of hyperbolic multinomial logistic regression:
\begin{multline}
v_k(x) = \\ \frac{2}{\sqrt{c}} \|z_k\| \sinh^{-1}\bigg(
\lambda^c_x \big\langle \sqrt{c}x, \frac{z_k}{\|z_k\|} \big\rangle \cosh(2\sqrt{c}r_k) \\
- (\lambda_x^c - 1) \sinh(2\sqrt{c}r_k)
\bigg)
\end{multline}
where $\big\langle .,. \big\rangle$ represents the Euclidean inner product. The final output of the Poincaré linear layer is  computed as:
\[
y = \frac{w}{1 + \sqrt{1 + c\|w\|}},
\]
where $w = (\frac{1}{\sqrt{c}}\sinh(\sqrt{c}v_k(x)))_{k=1}^m$.
\vspace{5mm}

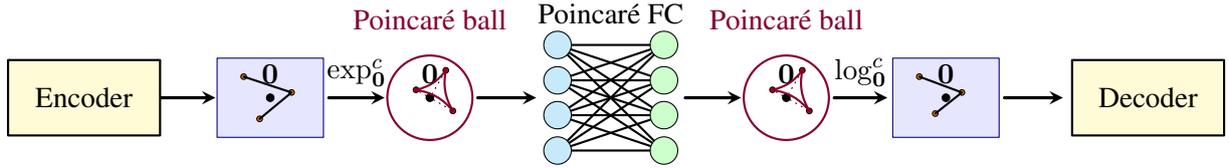
\begin{figure*}
  \centering
  \begin{tikzpicture}[scale=0.7, >=stealth, line cap=round, line join=round]

\begin{scope}[xshift=0.5cm]
  \node[draw=black, fill=yellow!20, thick,
        rectangle, minimum width=2.0cm, minimum height=1cm,
        align=center] (encoder) at (0,0) {Encoder};

  \coordinate (encoderEast) at (encoder.east);
\end{scope}

\begin{scope}[xshift=3.0cm]
  \coordinate (entryLeftPlane) at (-0.03,0);

  \path[fill=blue!10, draw=blue!50!black]
        (0,0.75) rectangle (2,-0.75);

  \coordinate (Oeuc) at (1,0);
  \node[draw=black,circle,fill=black,inner sep=1pt,
        label=above:$\mathbf{0}$] at (Oeuc) {};

  \coordinate (E1) at (0.5, 0.4);
  \coordinate (E2) at (0.8,-0.4);
  \coordinate (E3) at (1.4, 0.1);

  \foreach \pt in {E1,E2,E3} {
    \draw[fill=orange!90!black] (\pt) circle (1.5pt);
  }
  \draw[thick] (E1) -- (E3) -- (E2);

  \coordinate (leftPlaneEast) at (2.1,0);
\end{scope}

\draw[->, very thick] (encoderEast) -- (entryLeftPlane);

\begin{scope}[xshift=7.0cm]
  \coordinate (entryPoinc1) at (-0.85,0);

  \def\R{0.8}
  \draw[thick, purple!70!black] (0,0) circle (\R);

  \node[draw=black,circle,fill=black,inner sep=1pt,
        label=above:$\mathbf{0}$] (O1) at (0,0) {};

  \coordinate (P1) at ({0.6*cos(60)},{0.6*sin(60)});
  \coordinate (P2) at ({0.5*cos(-30)},{0.5*sin(-30)});
  \coordinate (P3) at ({0.3*cos(150)},{0.3*sin(150)});

  \foreach \pt in {P1,P2,P3} {
    \draw[fill=red!80!black] (\pt) circle (1.5pt);
  }

  \draw[dotted] (O1) -- (P1);
  \draw[dotted] (O1) -- (P2);
  \draw[dotted] (O1) -- (P3);

  \draw[thick, purple!70!black] (P1) to[bend right=20] (P2);
  \draw[thick, purple!70!black] (P2) to[bend right=20] (P3);
  \draw[thick, purple!70!black] (P3) to[bend right=20] (P1);

  \node[above, purple!70!black]
    at (0, {0.8+0.3})
    {Poincaré ball};

  \coordinate (poinc1East) at (0.9,0);
\end{scope}

\draw[->, very thick] (leftPlaneEast) -- (entryPoinc1)
  node[midway,above] {$\exp_{\mathbf{0}}^{c}$};

\begin{scope}[xshift=9.4cm]
  \coordinate (entryFC) at (-0.4,0);

  \foreach \index/\y in {1/1, 2/0.33, 3/-0.33, 4/-1} {
    \node[draw=black, circle, fill=cyan!20, minimum size=6pt]
          (in\index) at (0,\y) {};
  }

  \foreach \index/\y in {1/1, 2/0.33, 3/-0.33, 4/-1} {
    \node[draw=black, circle, fill=green!20, minimum size=6pt]
          (out\index) at (2,\y) {};
  }

  \foreach \i in {1,2,3,4} {
    \foreach \j in {1,2,3,4} {
      \draw[thick] (in\i) -- (out\j);
    }
  }

  \node[above] at (1,1.2) {Poincaré FC};

  \coordinate (fcEast) at (2.4,0);
\end{scope}

\draw[->, very thick] (poinc1East) -- (entryFC);

\begin{scope}[xshift=13.7cm]
  \coordinate (entryPoinc2) at (-0.85,0);

  \def\R{0.8}
  \draw[thick, purple!70!black] (0,0) circle (\R);
  \node[draw=black,circle,fill=black,inner sep=1pt,
        label=above:$\mathbf{0}$] (O2) at (0,0) {};

  \coordinate (Q1) at ({0.6*cos(60)},{0.6*sin(60)});
  \coordinate (Q2) at ({0.5*cos(-30)},{0.5*sin(-30)});
  \coordinate (Q3) at ({0.3*cos(150)},{0.3*sin(150)});

  \foreach \pt in {Q1,Q2,Q3} {
    \draw[fill=red!80!black] (\pt) circle (1.5pt);
  }
  \draw[dotted] (O2) -- (Q1);
  \draw[dotted] (O2) -- (Q2);
  \draw[dotted] (O2) -- (Q3);

  \draw[thick, purple!70!black] (Q1) to[bend right=20] (Q2);
  \draw[thick, purple!70!black] (Q2) to[bend right=20] (Q3);
  \draw[thick, purple!70!black] (Q3) to[bend right=20] (Q1);

  \node[above, purple!70!black]
    at (0, {0.8 + 0.3})
    {Poincaré ball};

  \coordinate (poinc2East) at (0.9,0);
\end{scope}

\draw[->, very thick] (fcEast) -- (entryPoinc2);

\begin{scope}[xshift=15.7cm]
  \coordinate (entryRightEuc) at (-0.1,0);

  \path[fill=blue!10, draw=blue!50!black]
       (0,0.75) rectangle (2,-0.75);

  \coordinate (OeucRight) at (1,0);
  \node[draw=black,circle,fill=black,inner sep=1pt,
        label=above:$\mathbf{0}$] at (OeucRight) {};

  \coordinate (E4) at (0.5,0.4);
  \coordinate (E5) at (0.8,-0.3);
  \coordinate (E6) at (1.3,0.1);

  \foreach \pt in {E4,E5,E6} {
    \draw[fill=orange!90!black] (\pt) circle (1.5pt);
  }
  \draw[thick] (E4) -- (E6) -- (E5);

  \coordinate (rightEucEast) at (2.1,0);
\end{scope}

\draw[->, very thick] (poinc2East) -- (entryRightEuc)
  node[midway,above] {$\log_{\mathbf{0}}^{c}$};

\begin{scope}[xshift=20.5cm]
  \coordinate (entryDecoder) at (-1.6,0);

  \node[draw=black, fill=yellow!20, thick,
        rectangle, minimum width=2.0cm, minimum height=1cm,
        align=center] (decoder) at (0,0) {Decoder};
\end{scope}

\draw[->, very thick] (rightEucEast) -- (entryDecoder);

\end{tikzpicture}
  \caption{Model architecture for our approach. After the T5 encoder, we project the resulting Euclidean embeddings onto the Poincaré ball. These hyperbolic embeddings are then refined through a trainable Poincaré layer. For compatibility with the T5 decoder, we project the hyperbolic embeddings back to the Euclidean space. While the T5 encoder and decoder parameters remain frozen throughout the training, the input contains trainable soft prompts.}
  \label{fig:model_architecture}
\end{figure*}

\subsection{Delta Hyperbolicity}\label{deltahyperbolicity}
Delta hyperbolicity ($\delta$-hyperbolicity) quantifies the extent to which a space is similar to a tree. This property makes it particularly relevant for analyzing and optimizing the curvature of multi-hop reasoning datasets. 

\noindent
\textbf{Gromov product.} The Gromov product is defined for points $x,y,w$ in a metric space $(X, d)$ as:
\[
(x,y)_w = \frac{1}{2} [d(x,w)+d(y,w) - d(x,y)].
\]
A metric space is $\delta$-hyperbolic if, for any four points $w,x,y,z \in X$, the inequality:
\[
(x,z)_w \geq \min\{(x,y)_w, (y,z)_w\} - \delta
\]
is satisfied. Smaller $\delta$ values indicate a closer resemblance to a tree-like structure. 
In our work, we adopt the approach outlined in \citet{khrulkov2020hyperbolicimageembeddings, sawhney-etal-2024-adapt, ermolov2022hyperbolicvisiontransformerscombining}, using $\delta$-hyperbolicity as a scale-invariant measure to assess the hyperbolic nature of the dataset. Specifically, we estimate the hyperbolicity constant $\delta(X)$, which represents the smallest possible $\delta$ satisfying the four-point condition for all quadruples of points in $X$. To account for variations in the scale of the dataset, we compute the relative hyperbolicity as
\[
\delta_{rel}(X) = \frac{2\delta(X)}{\text{diam}(X)},
\]
where diam($X$) represents the diameter of the dataset, defined as the maximum pairwise distance between points. Since $\delta_{rel}(X)$ is normalized by the diameter of the dataset, it remains invariant under uniform rescaling of distances, ensuring comparability across datasets of different scales. By construction $\delta_{rel}(X) \in [0,1]$, with values closer to zero indicating a strong resemblance to hyperbolic spaces.
Using the estimated $\delta_{rel}(X)$, we compute the curvature $c(X)$ of the embedding space following the formula provided by \citet{khrulkov2020hyperbolicimageembeddings}:
\begin{equation}\label{curvature_formula}
    c(X) = (\frac{0.144}{\delta_{rel}(X)})^2
\end{equation}
This calculation enables us to determine the curvature hyperparameter $c$.

\section{Method}

Our approach builds on the PaTH method \citet{misra2023triggeringmultihopreasoningquestion}, a two-step framework that fine-tunes the T5 model using soft prompts, added as trainable parameters to the input embeddings.
\subsection{PaTH Method Overview}
The PaTH method involves two primary stages of \textbf{knowledge integration} and \textbf{soft prompt tuning}. 
First, the T5 model is fine-tuned on the knowledge graph using the triples of the entity-relation-entity form $(e_1, r_1, e_2)$, enabling the model to internalize the foundational entity-relation structures. For each dataset, we only use the subgraph of triples relevant to our 2-hop questions (e.g., the paths connecting entities in each question) similar to \citet{misra2023triggeringmultihopreasoningquestion}.
In soft prompt tuning, two distinct soft prompts, called \textbf{parsing prompt} and \textbf{hopping prompt}, are trained to facilitate question parsing and reasoning tasks. The parsing prompt parses a question into an incomplete sequence $(e_1, r_1, r_2, ..., r_n)$, which serves as the input for the hopping prompt in the reasoning step.
The hopping prompt is trained using uniform random walks over the knowledge graph. The walks from the dev and test sets are excluded. Given an incomplete sequence representing the starting entity and intermediate relations $(e_1, r_1, r_2, ..., r_n)$, the model is tasked with predicting the complete sequence, including the intermediate entities and relations $(e_1, r_1, e_2, r_2, ..., r_{n-1}, e_n)$. This enables the model to infer reasoning paths by referring to the incomplete path.

\subsection{Incorporating Hyperbolic Representations}
Figure~\ref{fig:model_architecture} illustrates our simple integration of a hyperbolic layer into the T5 model.

The Euclidean embeddings generated by the T5 encoder are mapped onto the Poincaré ball using exponential mapping. 
Then they are processed through the hyperbolic layer, specifically designed for operations within the Poincaré space to preserve their geometric properties. After the transformation, the embeddings are mapped back to the Euclidean space using logarithmic mapping. This step enables compatibility with the T5 decoder for effective downstream processing. For the hyperbolic operations and Poincaré layer, we use the open-sourced implementation given by \citet{vanspengler2023poincareresnet}. 

\section{Experimental Setup}

For all experiments, we used the T5-Large model (770M parameters) \cite{raffel2023exploringlimitstransferlearning}. This model was fine-tuned using checkpoints adapted through the prefix LM objective \cite{liu2018generating} over 100,000 steps. We adopt the hyperparameters presented in \citet{misra2023triggeringmultihopreasoningquestion}, with a modification to the batch size, reducing it to 64 to accommodate hardware limitations. This adjustment applies to both the knowledge integration and prompt tuning processes. The optimizer is AdaFactor \cite{shazeer2018adafactor} and for the additional hyperbolic layer, we use the same learning rate of 0.001 used to fine-tune the T5 model. 

The curvature $c$ is initialized using Formula \ref{curvature_formula} in Section \ref{deltahyperbolicity}, which is based on the $\delta$-hyperbolicity of the dataset. 

Since computing $\delta$-hyperbolicity can be computationally expensive we calculate it in batches. We sample $1500$ points from the training dataset and compute $\delta_{rel}$. We repeat this process 5 times. For evaluation, we use the codebase of \citet{ho2020multihopqa}, which is open-sourced\footnote{\href{https://github.com/Alab-NII/2wikimultihop}{https://github.com/Alab-NII/2wikimultihop}}. Similarly to \citet{misra2023triggeringmultihopreasoningquestion}, we evaluate the model performance with the Exact Match (EM) score. 

\begin{table}[t]
\setlength{\tabcolsep}{3.5pt}
\small
\centering
\begin{tabularx}{\columnwidth}{Xccc}
\toprule
Dataset & Nodes & Edges & Relations  \\ 
\midrule
2Wiki\-MultiHopQA & 97,298 & 95,116 & 29 \\
MetaQA & 31,374 & 58,974 & 9 \\
MLPQ & 51,402 & 53,327 & 72 \\
PQ & 1,056 & 1,211 & 13 \\ 
\bottomrule
\end{tabularx}
\caption{Knowledge graph statistics of the datasets}
\label{tab:knowledge_graph_statistics}
\end{table}

\subsection{Dataset Preparation} \label{sec:datasets}
We use four datasets in a closed-book QA setting, where context was omitted to prioritize the reasoning capabilities of the model. The complete statistics for these datasets can be found in Tables \ref{tab:knowledge_graph_statistics} and \ref{tab:num_questions}. To ensure consistent evaluation across all datasets, we focus on the 2-hop questions. 

\noindent
\textbf{2WikiMultiHopQA} 

\cite{ho2020multihopqa}, hereafter referred to as 2WikiHop for simplicity, consists of two-hop English questions constructed over a knowledge base containing 98,284 entities and 29 relations sourced from WikiData \cite{vrandevcic2014wikidata}. 
 
Since the test splits of the 2WikiHop are private, the validation split was repurposed as the test set, with 10\% of the training data reserved for validation. This adaptation mirrors the approach taken by \citet{misra2023triggeringmultihopreasoningquestion}.

\begin{table}[t]
\setlength{\tabcolsep}{3.5pt}
\centering
\small
\begin{tabularx}{\columnwidth}{Xccc}
\toprule
Dataset & Train & Dev & Test  \\ 
\midrule
2Wiki\-MultiHopQA & 72,760 & 8,085 & 6,768 \\
MetaQA & 47,108 & 5,951 & 5,942 \\
MLPQ & 57,283 & 7,160 & 7,161 \\
PQ & 1,698 & 210 & 191 \\
\bottomrule
\end{tabularx}
\caption{Number of questions in train/dev/test splits.}
\label{tab:num_questions}
\end{table}

\noindent
\textbf{MetaQA} consists of questions that can have multiple answers, different from 2WikiHop, where each question is associated with a single answer. For our study, we focused exclusively on a subset of the MetaQA dataset containing questions with a single possible answer to ensure consistency in the evaluation.
In particular, MetaQA does not directly provide evidence for each question. To address this, we generated the necessary evidence for each question, as detailed in Appendix \ref{sec:evidence_creation}. For this dataset, we used the official train/dev/test split\footnote{\href{https://github.com/yuyuz/MetaQA}{https://github.com/yuyuz/MetaQA}}.

\noindent
\textbf{MLPQ} \cite{mlpq2023} consists of multilingual questions paired with corresponding language-specific knowledge graphs. For our study, we use the evidence (paths) of the dataset as a unified knowledge graph, resulting in a total of 51,401 entities and 72 relations. We specifically focus on files that contain English questions alongside potential French-language entities and relations. To maintain consistency during question parsing, French relations are translated into English. The dataset is divided into training, validation, and test sets with a ratio of 8:1:1, resulting in 57,283 questions for training, 7,160 for validation, and 7,161 for testing. Although the evidence parts are structured as triples, it is worth noting that due to the multilingual nature of the dataset, the tail of the first evidence may not always match the head of the subsequent evidence. To address this, we normalize by always selecting the English entity to construct the knowledge graph.

\noindent
The Path Questions (\textbf{PQ}) dataset \cite{zhou2018interpretable} is a QA dataset designed for multi-hop reasoning, leveraging entity relationships sourced from a knowledge base called Freebase \cite{bollacker2008freebase}. Our focus is on the 2-hop reasoning subset, which comprises 1,908 questions, their corresponding answers, and the reasoning paths used to derive them. We adopt the same dataset split as \citet{wang2024hypermr}, which is an 8:1:1 ratio for training, dev, and test sets, respectively. However, contrary to \citet{wang2024hypermr}, we exclude the reasoning walks found in the dev and test splits during training, making the task more challenging since all supporting evidence present in the dev and test sets was also part of their training split. 

We chose PQ over the similar PQL~\cite{zhou-etal-2018-interpretable} dataset since PQ’s smaller size allowed us to stay within our computational budget while capturing the same multi-hop reasoning patterns.

\section{Results}
In this section, we compare the two spaces in a variety of settings, provide an ablation study on the curvature, show that computationally both cases are similar, present the results of distance analysis in the two spaces, and give some insights into dataset difficulty.

\begin{table}
\centering
\small
\setlength{\tabcolsep}{3.0pt}
\begin{tabularx}{\columnwidth}{Xlcccc}
\toprule
Data&Model &2WikiHop & MetaQA & MLPQ & PQ \\ 
\midrule 
Dev&Euclidean & 44.36 & 22.92 & 81.03 & 18.28 \\ 
Dev&Hyperbolic & \textbf{46.93} & \textbf{28.33} & \textbf{82.60} & \textbf{29.03} \\ 
\midrule
Test&Euclidean & 14.88 & 19.76 & 72.10 & 11.90\\ 
Test&Hyperbolic & \textbf{15.20} & \textbf{25.40} & \textbf{74.58} & \textbf{23.21} \\ 
\bottomrule
\end{tabularx}
\caption{Exact match scores for hopping prompt}
\label{tab:performance_hopping_val}
\end{table}

\begin{table}
\centering
\small
\setlength{\tabcolsep}{3.0pt}
\begin{tabularx}{\columnwidth}{Xlcccc}
\toprule
Data & Model & 2WikiHop & MetaQA & MLPQ & PQ \\ 
\midrule
Dev&Euclidean & 88.60 & 95.51 & 97.08 & \textbf{100} \\ 
Dev&Hyperbolic & \textbf{89.34} & \textbf{95.65} & \textbf{97.14} & \textbf{100} \\ 
\midrule
Test&Euclidean & 79.24 & \textbf{95.27} & 95.91 & \textbf{98.95}\\ 
Test&Hyperbolic & \textbf{80.11} & 95.07 & \textbf{96.79} & \textbf{98.95}\\ 
\bottomrule
\end{tabularx}
\caption{Exact match scores for parsing prompt}
\label{tab:performance_parsing_val}
\end{table}

\subsection{Hyberbolic vs. Euclidean Layer} 
Table \ref{tab:performance_hopping_val} presents the results across all datasets under prompt tuning for the hopping prompt. The results demonstrate that the hyperbolic layer outperforms the Euclidean counterpart across all datasets. For 2WikiHop, the hyperbolic layer increases the EM score from 44.36\% (Euclidean) to 46.93\%, reflecting a performance boost of 2.57\%. Similarly, for MetaQA, the hyperbolic layer achieves an exact match (EM) score of 28.33\%, outperforming the Euclidean layer, which achieves 22.92\%, resulting in an improvement of 5.41\%.

\begin{table}
\centering
\small
\setlength{\tabcolsep}{1.5pt}
\begin{tabularx}{\columnwidth}{llcccr}
\toprule
Parsing & Hopping & 2WikiHop & MetaQA & MLPQ & PQ \\ \midrule
Euclidean & Euclidean & 13.39 & 19.20 & 72.59 & 12.04 \\
Hyperbolic & Euclidean & 13.56 & 19.08 & 72.74 & 12.04 \\
Euclidean & Hyperbolic & 13.40 & \textbf{24.74} &  \textbf{73.48} & \textbf{23.04} \\ 
Hyperbolic & Hyperbolic & \textbf{13.65} & 24.72 & 73.40 & 22.51 \\ \bottomrule
\end{tabularx}
\caption{Exact match scores on test set for T5 with the additional Euclidean/hyperbolic layer for both prompts.}
\label{tab:performance_test}
\end{table}

Interestingly, the smallest improvement occurs in MLPQ, where the hyperbolic layer increases the EM score from 81.03\% to 82.60\%, marking a marginal gain of 1.57\%. This limited improvement could be attributed to the already high baseline performance achieved by the Euclidean layer in this dataset. With less room for improvement, the hierarchical modeling advantages of the hyperbolic layer are less pronounced. Notably, MLPQ’s predominantly linear knowledge‐graph structure—with over 80\% of nodes having an out‐degree of one (see Figure~\ref{fig:out_degree_comparison})—constrains the benefits of hyperbolic space on this dataset. For the same reason, we can also see marginal improvements for the parsing prompt in Table \ref{tab:performance_parsing_val}.
This contrasts with datasets like MetaQA and PQ (for the hopping prompt in Table \ref{tab:performance_hopping_val}), where a lower baseline provides more opportunities for substantial gains. More importantly, given that the parsing task is independent of the knowledge graph structure, it might not inherently benefit from the hierarchical properties of the hyperbolic space.  

\begin{table}
\centering
\small
\setlength{\tabcolsep}{3.0pt}
\begin{tabularx}{\columnwidth}{Xlcccc}
\toprule
Data & Model & $\text{2WikiHop}$ & $\text{MetaQA}$ & $\text{MLPQ}$ & $\text{PQ}$\\ 
\midrule
Dev & Euclidean & 32.43 & 13.89 & 78.74 & 12.76 \\ 
Dev & Hyperbolic & \textbf{34.22} & \textbf{17.09} & \textbf{79.64} & \textbf{27.42} \\ 
\midrule
Test & Euclidean & 10.24 & 12.08 & 70.36 & 11.90 \\
Test & Hyperbolic & \textbf{10.70} & \textbf{15.10} & \textbf{72.12} & \textbf{22.02} \\
\bottomrule
\end{tabularx}
\caption{Exact match scores for hopping stage when applying the additional layer without soft prompts.}
\label{tab:validation_performance_no_soft_prompt}
\end{table}

Table \ref{tab:performance_test} compares all configurations of hyperbolic and Euclidean layers for the parsing and hopping stages. The results show that in most cases the use of hyperbolic space for the hopping stage gives the highest performance improvements. This finding is expected as the hopping stage is influenced by the knowledge graph hierarchies, which are better captured in hyperbolic space. Notably, while the Euclidean-hyperbolic configuration for parsing-hopping stages achieves the best performance in most cases, the hyperbolic-hyperbolic configuration follows closely, with only a marginal difference (ranging from 0.02 to 0.47 EM points). This suggests that while hyperbolic hopping significantly improves the performance, the choice of space for the parsing stage could also have a relatively small impact. 

Table \ref{tab:validation_performance_no_soft_prompt} presents the results of the additional Euclidean and hyperbolic layers on the random walk dev set without the use of soft prompts. The results demonstrate that the hyperbolic layer consistently outperforms the Euclidean layer, achieving higher scores on all datasets. This indicates that the performance observed is not dependent on soft prompting, as the hyperbolic layer exhibits superior results even in its absence. 

\begin{figure}
    \centering
    \includegraphics[width=\linewidth]{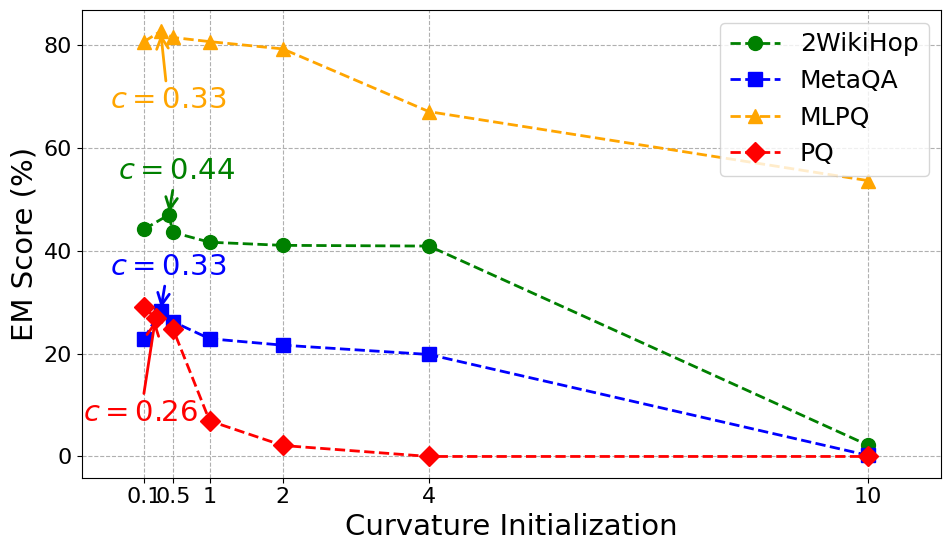}
    \caption{Curvature ablation for random walk training showing exact match score on dev sets. Initializing the curvature with or around $\delta$-hyperbolicity gives the highest EM score.}
    \label{fig:curvature_ablation}
\end{figure}

\begin{table}[t]
\small  
\centering
\setlength{\tabcolsep}{3.5pt}
\begin{tabularx}{\columnwidth}{X |c |c |c |c}
\toprule
 & \textbf{2WikiHop} & \textbf{MetaQA} & \textbf{MLPQ} & \textbf{PQ} \\ 
\midrule
\multicolumn{5}{c}{\textit{Random Walk Dataset (Hopping Prompt)}} \\ 
\midrule
$\delta$ & 0.22\tiny{$\pm$0.012} & 0.25\tiny{$\pm$0.015} & 0.25\tiny{$\pm$0.013} & 0.28\tiny{$\pm$0.017} \\ 
$c$      & 0.44          & 0.33          & 0.33 &  0.26          \\ 
\midrule
\multicolumn{5}{c}{\textit{Parsing Dataset (Parsing Prompt)}} \\ 
\midrule
$\delta$ & 0.29\tiny{$\pm$0.017} & 0.33\tiny{$\pm$0.018} & 0.29\tiny{$\pm$0.020} & 0.28\tiny{$\pm$0.019} \\ 
$c$      & 0.25           & 0.19          & 0.25 & 0.26          \\ 
\bottomrule
\end{tabularx}
\caption{Mean $\delta$-hyperbolicity and curvature values for random walks and parsing prompt data.}
\label{tab:delta_hyperbolicities_combined}
\end{table}

\subsection{Curvature Ablation}
\label{sec:curvature_ablation}
One hyperparameter of the hyperbolic layer is the curvature, which can be initialized arbitrarily. Figure \ref{fig:curvature_ablation} presents the results of the curvature ablation study with different initializations. A key takeaway from this study is that initialization plays a crucial role in model performance. Specifically, setting the curvature based on the relative $\delta$-hyperbolicity (see Section \ref{deltahyperbolicity}) of each dataset, as shown in Table~\ref{tab:delta_hyperbolicities_combined}, yields the best (or very close to it for PQ) across all datasets. In contrast, while smaller curvatures such as $0.1$ and $1.0$ still yield competitive results, increasing the curvature beyond $1.0$ leads to a notable degradation in EM scores. For instance, with a curvature of $10.0$, the EM scores drop drastically to $2.21\%$ for 2WikiHop and $0.22\%$ for MetaQA, demonstrating that inappropriate curvature values can severely impact model effectiveness. These findings suggest that hyperbolic models benefit from curvature settings that reflect the structure of the data. Since hyperbolic space expands exponentially, setting the curvature to match a dataset’s $\delta$-hyperbolicity allows the model to better reflect hierarchical relationships, thereby improving multi-hop reasoning accuracy. In contrast, Euclidean space lacks this adaptability, making it less effective when reasoning over complex data in knowledge graphs.

\subsection{Computational Analysis}
\begin{figure}[t]
    \centering
    \includegraphics[width=0.8\linewidth]{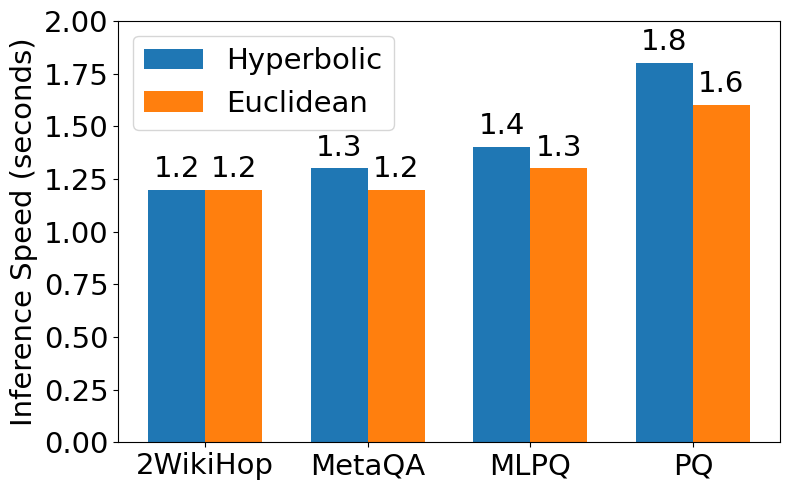}
    \caption{Average inference time per batch (=8) on the test data. The hyperbolic layer causes a negligible increase in inference time over the Euclidean layer.}
    \label{fig:inference_speed_comparison}
\end{figure}

Another crucial aspect of this study is the computational complexity associated with hyperbolic layers. Despite improved performance, adding a hyperbolic layer introduces negligible time and memory overhead as shown in Figure~\ref{fig:inference_speed_comparison}. This observation is significant because it demonstrates that hyperbolic layers can achieve superior performance without significantly increasing the computational cost of the model. This makes hyperbolic layers a practical and efficient choice for tasks involving graph-structured data, such as multi-hop reasoning. 

\begin{table}[t]
\centering
\small
\setlength{\tabcolsep}{3pt}
\begin{tabularx}{\columnwidth}{Xcccc}
\toprule
Hopping & 2WikiHop & MetaQA & MLPQ & PQ \\ 
\midrule
First relation & 100 & 100 & 81.25 & 100 \\ 
Second relation & 100 & 100 & 69.01 & 99.46 \\ 
\bottomrule
\end{tabularx}
\caption{Percentage of cases where the geodesic distance in hyperbolic space between the source entity and its relations is larger than the Euclidean distance.}
\label{tab:distance_relation_comparison}
\end{table} 

\subsection{Embedding Distances}
\label{sec:embedding_distances}
To investigate how the distance between the source entities and their relations compare to each other in the Euclidean and hyperbolic layers, we looked into their embeddings in these layers. Table \ref{tab:distance_relation_comparison} presents a comparison between hyperbolic and Euclidean embeddings in terms of their geodesic and Euclidean distances. Specifically, it reports the percentage of cases where the geodesic distance in hyperbolic space is larger than the Euclidean distance for the Euclidean embeddings. The comparison is conducted for both the first and second relational hops with respect to the source entity.

\begin{figure}
    \centering
    \includegraphics[width=1\linewidth]{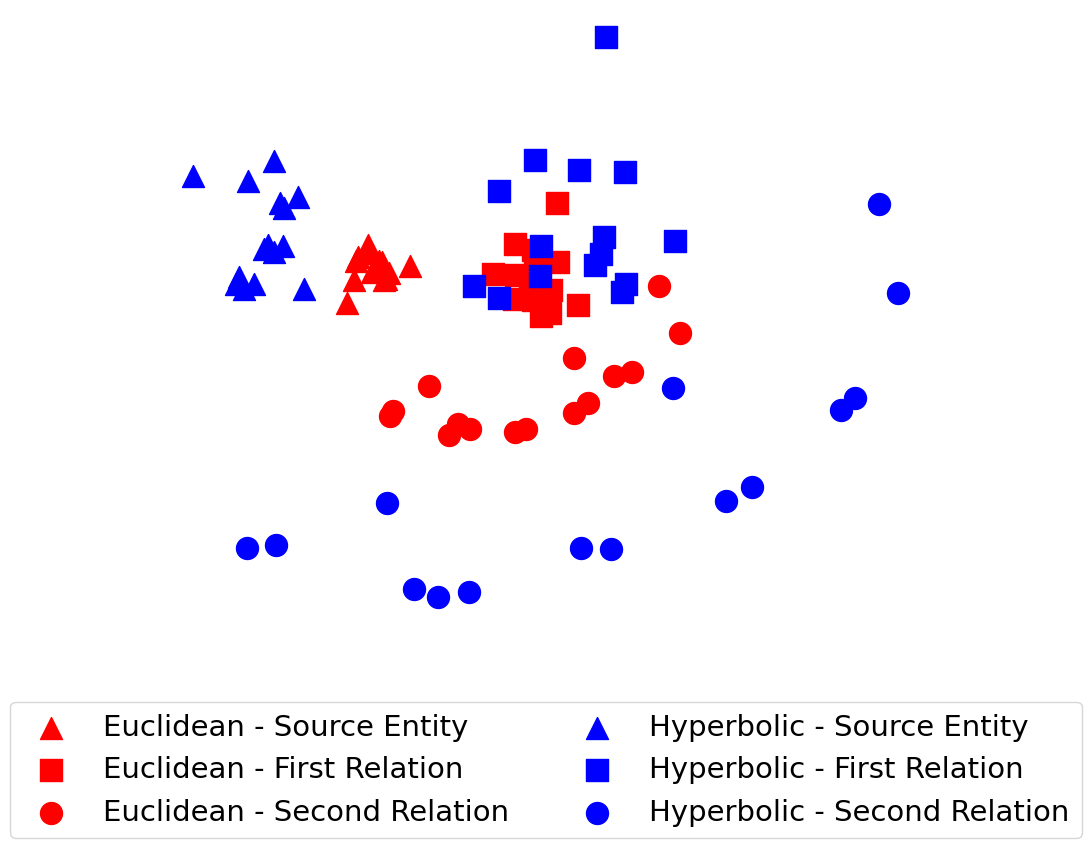}
    \caption{Embeddings for 15 input samples from the MetaQA dataset, each structured as "source entity; first relation; second relation" in Euclidean versus Poincaré layer. The Euclidean embeddings use Euclidean distance while the hyperbolic embeddings use geodesic distance. Due to the exponential growth of the hyperbolic space, entities and relations can be more spread out as the paths become longer.}
    \label{fig:embedding_distances}
\end{figure}

\begin{figure}
    \centering
    \includegraphics[width=0.95\linewidth]{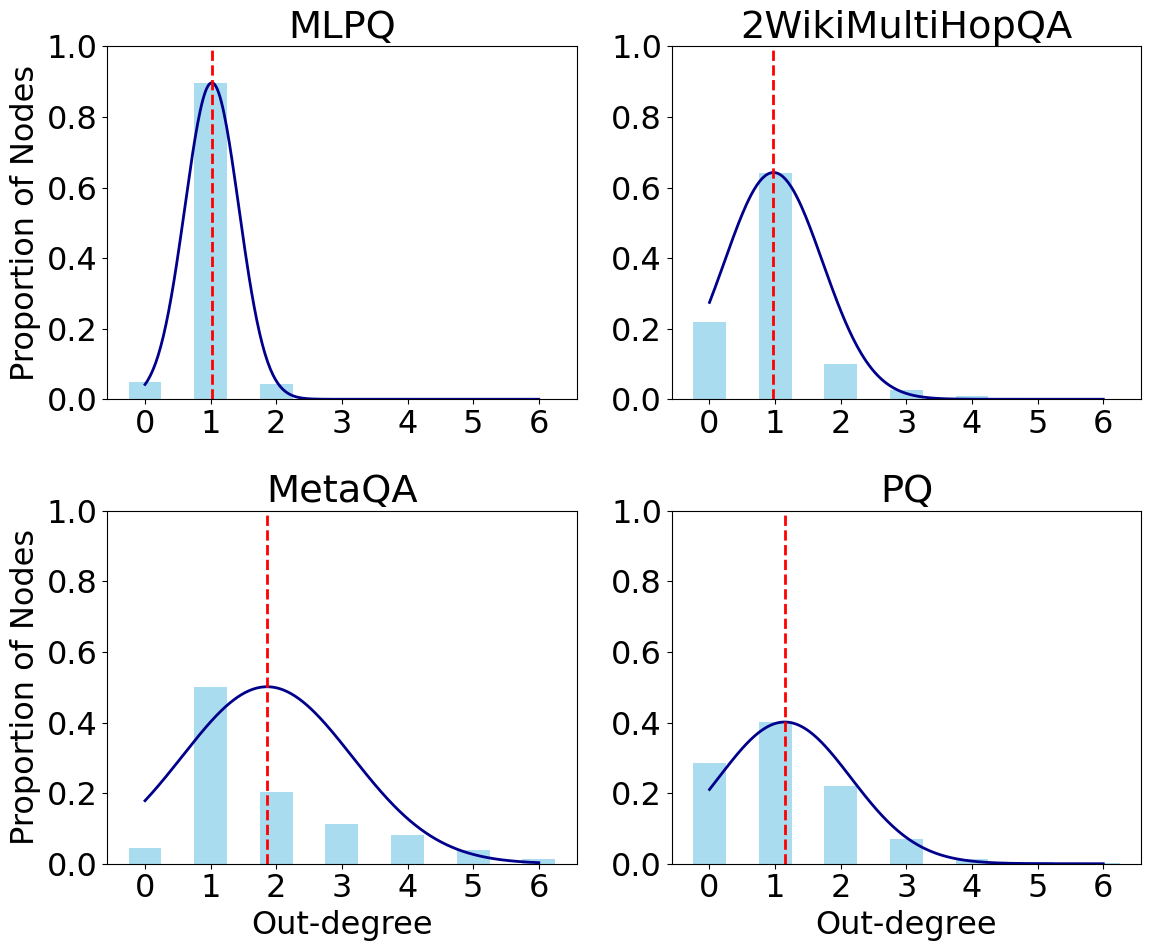}
    \caption{Distribution of out-going relations (out-degree) for each dataset. As the proportion of nodes with a degree of 2 or higher goes up, the complexity of the dataset also increases. }
    \label{fig:out_degree_comparison}
\end{figure}

For 2WikiHop, MetaQA, and PQ datasets, it is evident that the hyperbolic relation embeddings consistently exhibit a greater distance from the source entity in comparison to the Euclidean embeddings, as evidenced by almost 100\% of the cases. However, MLPQ demonstrates a notable decrease in these percentages for both the first (81.25\%) and the second (69.01\%) relational hops. This behavior can be attributed to the structural characteristics inherent in the MLPQ knowledge graph. The fact that over 80\% of its nodes have an out-degree of 1 (only one out-going relation as seen in Figure \ref{fig:out_degree_comparison}) indicates that MLPQ is predominantly a linear knowledge graph rather than a hierarchical one. The hyperbolic space is particularly beneficial for tree-like structures, where distances expand exponentially with branching. However, in a mostly linear graph, entities and relations are more evenly spaced out, meaning that Euclidean space can capture these relationships almost just as effectively. Since MLPQ lacks significant tree-like expansion, its hyperbolic distances are not consistently larger than Euclidean distances, leading to significantly lower percentages compared to the other datasets. 

The results confirm that for tree-like knowledge graphs, the hyperbolic geodesic distance typically exceeds the Euclidean distance. This outcome is expected due to the exponential expansion of hyperbolic space. Unlike Euclidean space, where distance scales linearly, hyperbolic space exhibits exponential growth, allowing entities and relations to be more sparsely distributed (Figure~\ref{fig:embedding_distances}), which, in turn, makes it easier for the model to find relevant paths. 

Such geometric properties yield a better theoretical justification for the performance gains seen in hyperbolic models. By allowing for increased spatial separation between entities along a path, hyperbolic space lessens interference between rival paths and improves the model’s capability to learn effective reasoning multi-hop chains. This has a significant impact in scenarios where disambiguation between relation paths proves central—something Euclidean space has difficulty with given its linear growth and reduced ability to represent hierarchical branching. Hence, the benefit is not simply in numeric improvements, but rather in how the geometry changes the landscape of the embedding to better reflect the nature of reasoning problems.

\subsection{Dataset Difficulty}
Depending on the knowledge graph, datasets can have different levels of difficulty. Figure \ref{fig:out_degree_comparison} presents the proportion of nodes in each dataset with their out-going relations (out-degree) in their knowledge graphs. The MLPQ dataset is comparatively simpler, with over 80\% of its nodes having an out-degree of 1. This characteristic significantly reduces the complexity of navigating the graph, as there is typically only one possible path from a given source node. In contrast, the MetaQA is more challenging, with only 50\% of its nodes having an out-degree of 1 while more than 40\% possess an out-degree of 2 or higher. The presence of multiple paths increases ambiguity, making traversal more complex and negatively impacting the performance, particularly in the random walk stage.

\section{Conclusion and Future Work}

We carried out a rigorous and careful investigation of using hyperbolic versus Euclidean representations in multi-hop reasoning and showed some of the advantages of the former compared to the latter. 

Our experiments also confirm that initializing the curvature using the relative delta hyperbolicity of the dataset provides a robust and effective starting point for learning, ensuring that the model captures the hierarchical relationships within the data with greater accuracy. We also provided evidence for the hyperbolic geometry showing more effectiveness when the dataset has more hierarchical characteristics. These findings underscore the importance of understanding the structural properties of the data when selecting appropriate model architectures. In addition, our findings open several promising avenues for future research. 

Given that the number of outgoing relations for each node in different knowledge graphs is not equal in many cases, future work could investigate a general manifold structure as well as other hyperbolic spaces for multi-hop reasoning. 

While our current framework focuses on encoder-decoder models in a closed-book QA setting, future work should investigate the generalizability of hyperbolic representations across broader architectures and tasks. First, extending our approach to decoder-only language models would help assess the geometric advantages in other generative models. Second, applying hyperbolic reasoning layers to open-book QA tasks, where models retrieve and integrate external evidence, would clarify the interaction between geometric embedding space and retrieval-based reasoning. Finally, examining our method on a wider variety of datasets and QA formats—including multi-lingual, noisy, or longer-hop reasoning datasets—will be critical to understanding the full scope of its effectiveness and limitations.

\section{Limitations}
Although our approach shows promising results, it has certain limitations: 
First, we focus exclusively on the closed-book QA setting, where no external context is provided to the model. This limitation inherently limits the amount of information available to answer questions, as the model relies solely on its trained knowledge. As a result, our approach may underperform compared to models that use additional context, such as open-book \cite{jiang2022understanding, feng2020scalable, xu2021exploiting} or retrieval-augmented \cite{yucheng-etal-2024, shi-etal-2024-generate, zhang2024hierarchical} methods, which can provide more relevant information during inference.
Second, our experiments were conducted using a frozen model, where only a small number of parameters in the additional layer were fine-tuned. While this approach reduces computational cost and maintains efficiency, it may limit the ability of the models that require full fine-tuning for higher accuracy. In such cases, the impact of one hyperbolic layer with only one million trainable parameters might fade away compared to a billion parameters.
\section{Ethics Statement}
In this study, we exclusively employ pre-trained knowledge from the T5 model and the datasets utilized in our experimental setup. We have solely transformed the data as outlined in Section~\ref{sec:datasets}, without introducing or curating additional external knowledge sources. However, it is crucial to acknowledge that the datasets may contain biased, inaccurate, or incomplete information, which could influence the model's reasoning and outputs. Furthermore, these datasets may inadvertently include private or sensitive information that has not been explicitly identified, as addressing such concerns is beyond the scope of this study.

\section*{Acknowledgments}
This work was partially supported by the AISafety Project, funded by the BMBF under the grant proposal 05D23PD1, and by the state of North Rhine-Westphalia as part of the Lamarr Institute for Machine Learning and Artificial Intelligence. 
We would like to thank Prof. Zorah Lähner and Dr. Benedikt Kolbe for their amazing insights during this work. We would also like to thank the reviewers for their invaluable comments, which helped strengthen the quality of this work.
\bibliography{anthology,custom}

\appendix

\section{Appendix}\label{sec:appendix}
\subsection{Evidence Creation for MetaQA}\label{sec:evidence_creation}
The MetaQA dataset contains structured information about questions, including the source entity, the tail entity (answer), and the intermediate relations that connect them for each question. For instance, in the question:
\begin{center} \textit{"What are the languages spoken in the films directed by [Joel Zwick]?"} \end{center}
The source entity is \textit{Joel Zwick}, and the answer would be \textit{Greek}. Additionally, the dataset provides the intermediate relations forming the reasoning path from the source entity to the answer. For this example, the intermediate relations are represented as a path string:
\begin{center} 
\textit{director\_to\_movie\_to\_language}. 
\end{center}
To find the evidence, we first parse the path string into the pairs:
\begin{itemize}
    \item (director, movie)
    \item (movie, language)
\end{itemize}
Each pair represents a segment of the reasoning path. These pairs are then mapped to their corresponding relations in the knowledge graph using the mapping defined in Table \ref{tab:pair_to_relation_mapping}. For instance:
\begin{itemize}
    \item (director, movie) $\rightarrow$ directed\_by\_reversed
    \item (movie, language) $\rightarrow$ in\_language
\end{itemize}

Using the source entity, the intermediate relations, and the answer entity, we construct the complete entity-relation-entity-relation-entity chain for each question. This chain serves as the evidence for the reasoning process.

\begin{table}
\centering
\small
\begin{tabularx}{\columnwidth}{XX}
\toprule
\textbf{Pair} & \textbf{Relation}  \\ 
\midrule
(movie, language) & in\_language\\
(movie, year) & release\_year\\
(movie, writer) & written\_by\\
(movie, director)& directed\_by\\
(movie, genre)& has\_genre\\
(movie, actor)& starred\_actors\\

(language, movie)& in\_language\_reversed\\
(year, movie)& release\_year\_reversed\\
(writer, movie)& written\_by\_reversed\\
(director, movie)& directed\_by\_reversed\\
(genre, movie)& has\_genre\_reversed\\
(actor, movie)& starred\_actors\_reversed\\
\bottomrule
\end{tabularx}
\caption{Pair to relation mapping of the MetaQA dataset}
\label{tab:pair_to_relation_mapping}
\end{table}

\subsection{Further Performance Comparison and Model Efficiency on the PQ Dataset}

\begin{table}[H]
\centering
\small
\begin{tabularx}{\columnwidth}{Xr}
\toprule
 & EM Score  \\ 
\midrule
Ours with Euclidean & 89.01 \\
Ours with Hyperbolic & 94.24 \\ 
HyperMR \cite{wang2024hypermr} & \textbf{96.2}   \\ 
\bottomrule
\end{tabularx}
\caption{Comparison of our model with HyperMR on the PQ dataset.}
\label{tab:pq_test_comparison}
\end{table}

\begin{figure}[h]
    \centering
    \includegraphics[width=1\linewidth]{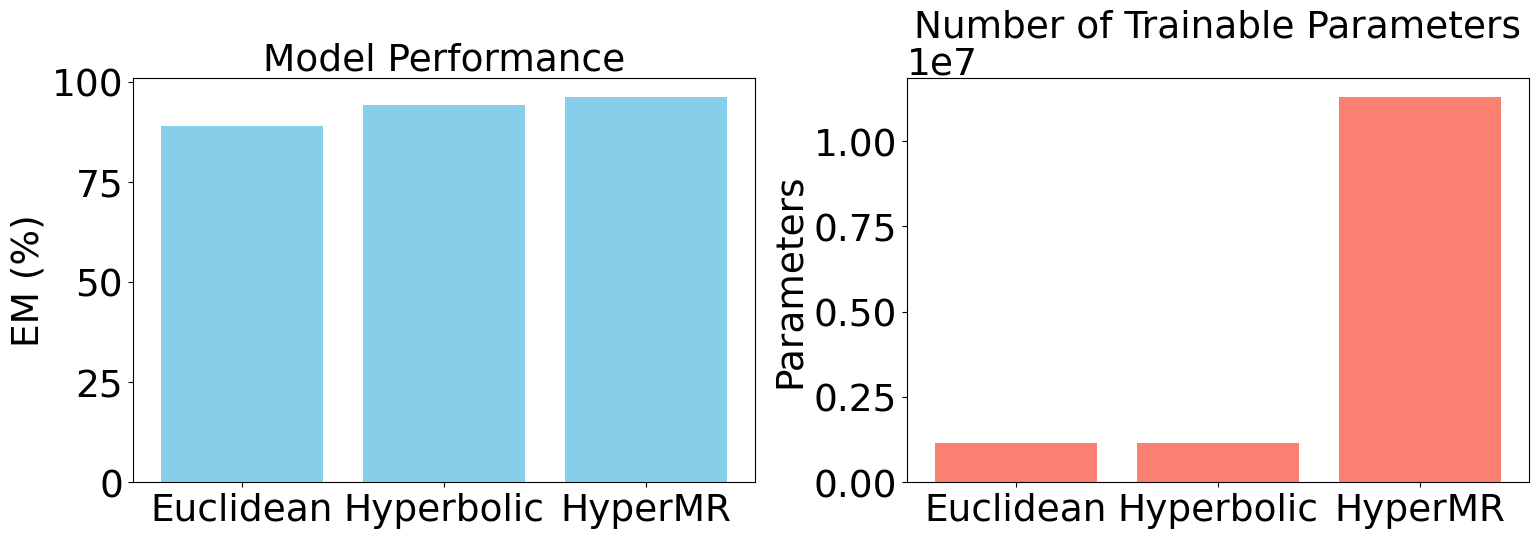}
    \caption{On the left: EM score on PQ test set for our model with Euclidean/hyperbolic layer vs HyperMR \citet{wang2024hypermr}. On the right: Number of trainable parameters. Even though our approach has lower performance, our model only has approximately $\frac{1}{10}$ of the trainable parameters as their model.}
    \label{fig:pq_trainable_parameters_comparison}
\end{figure}

In Table \ref{tab:pq_test_comparison}, we compare the test performance of our approach against the state-of-the-art hyperbolic model, HyperMR, on the PQ dataset. In this case, we incorporated reasoning paths into the training process to ensure a fair comparison. Our results indicate that the hyperbolic layer outperforms its Euclidean counterpart, improving accuracy from 89.01\% to 94.24\%. While our hyperbolic model performs slightly lower than HyperMR, it achieves this with only a fraction of the trainable parameters (one million compared to 11 million for HyperMR) as illustrated in Figure~\ref{fig:pq_trainable_parameters_comparison}.

\end{document}